\documentclass{article} 
\usepackage{iclr2027_conference,times}

\usepackage{amsmath,amsfonts,bm}

\def\eqref#1{equation~\ref{#1}}

\def\1{\bm{1}}

\DeclareMathAlphabet{\mathsfit}{\encodingdefault}{\sfdefault}{m}{sl}
\SetMathAlphabet{\mathsfit}{bold}{\encodingdefault}{\sfdefault}{bx}{n}

\usepackage{graphicx}
\usepackage{url}
\usepackage{booktabs}
\usepackage{longtable}
\usepackage{placeins}
\usepackage{hyperref}
\usepackage{multirow}
\usepackage{float}
\usepackage{wrapfig}
\usepackage{needspace}
\usepackage{amssymb}
\usepackage{capt-of}
\usepackage{tabularx}
\usepackage[table]{xcolor}
\usepackage{enumitem}

\definecolor{gainGreen}{HTML}{238443}
\definecolor{lossRed}{HTML}{C0392B}
\newcommand{\increaseGain}[1]{%
  \,{\scriptsize\color{gainGreen}\bfseries(+#1\%)}}

\newcommand{\decreaseGain}[1]{%
  \,{\scriptsize\color{gainGreen}\bfseries($-$#1\%)}}

\title{OctoNest: Adaptive Cross-Device Execution through Stateful Control}

\author{Shu Yao$^{1,2}$, Yuhua Luo$^1$, Qian Long$^1$, Jingru Fan$^1$, Yuheng Wang$^1$, Lin Wu$^1$, \\
\textbf{Zhuoyuan Yu$^1$, Yufan Dang$^3$, Huatao Li$^1$, Chen Qian$^1$\! \thanks{Corresponding author}}\\
$^1$School of Artificial Intelligence, Shanghai Jiao Tong University\\
$^2$Shanghai Innovation Institute \qquad $^3$Tsinghua University\\
yaoshu@sii.edu.cn \qquad qianc@sjtu.edu.cn\\[3pt]
{\normalfont\small\hypersetup{hidelinks}\textbf{Open-source code and benchmark:}~\href{https://github.com/AgenticOrgLab/OctoNest}{\texttt{github.com/AgenticOrgLab/OctoNest}}}}

\iclrfinalcopy 
\hypersetup{
  pdftitle={OctoNest: Adaptive Cross-Device Execution through Stateful Control},
  pdfauthor={Shu Yao, Yuhua Luo, Qian Long, Jingru Fan, Yuheng Wang, Lin Wu, Zhuoyuan Yu, Yufan Dang, Huatao Li, Chen Qian}
}

\makeatletter
\renewcommand{\@maketitle}{\vbox{\hsize\textwidth
{\noindent
  \makebox[0pt][r]{\raisebox{-6pt}{\includegraphics[height=0.30in]{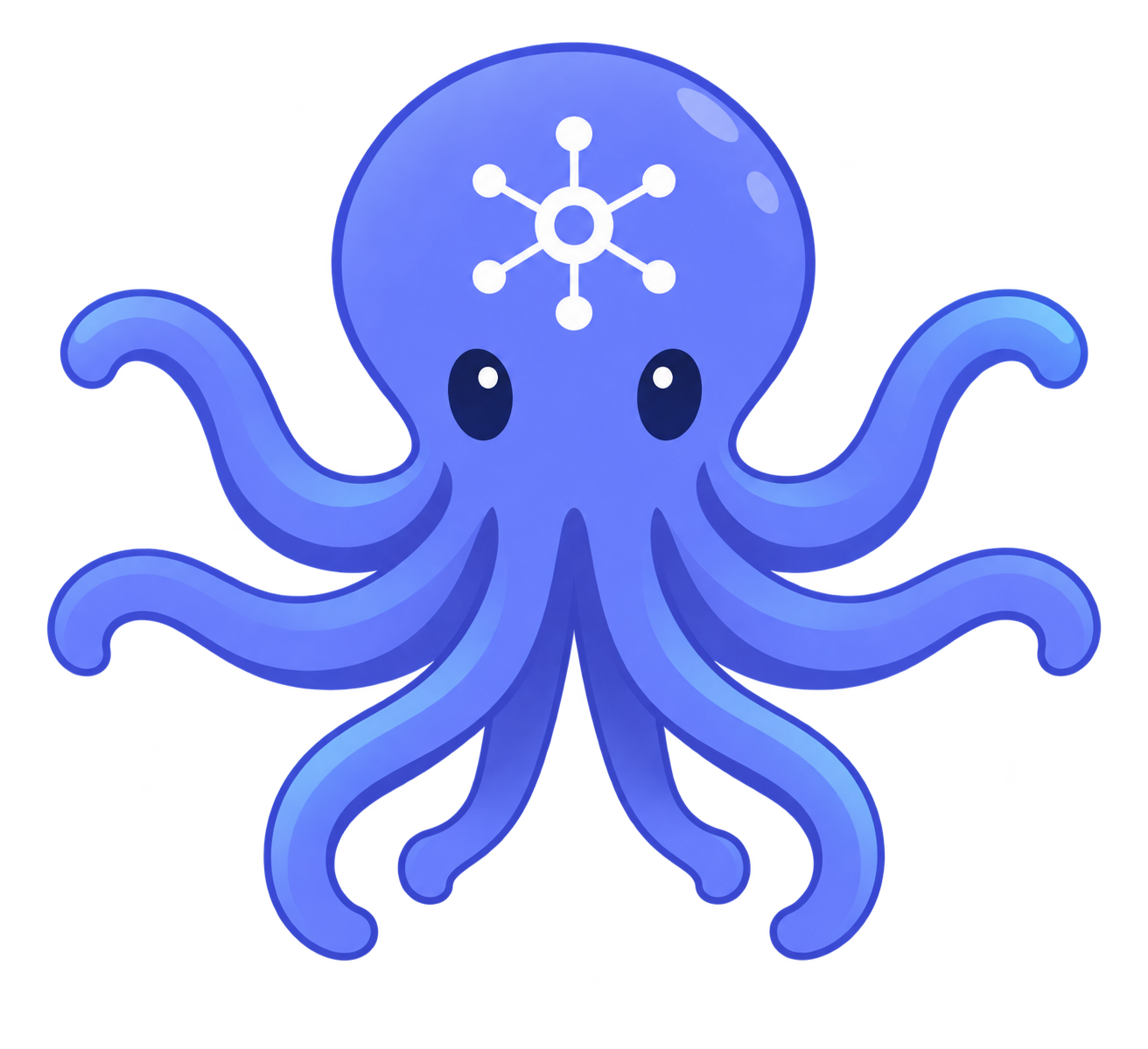}}\hspace{-8pt}}%
  \hspace{12pt}%
  \parbox[t]{\dimexpr\textwidth-12pt\relax}{\raggedright {\LARGE\sc \@title\par}}%
\par}
\ificlrfinal
    \lhead{}
    \def\And{\end{tabular}\hfil\linebreak[0]\hfil
            \begin{tabular}[t]{l}\bf\rule{\z@}{24pt}\ignorespaces}%
  \def\AND{\end{tabular}\hfil\linebreak[4]\hfil
            \begin{tabular}[t]{l}\bf\rule{\z@}{24pt}\ignorespaces}%
    \begin{tabular}[t]{l}\bf\rule{\z@}{24pt}\@author\end{tabular}%
\else
       \lhead{Under review as a conference paper at ICLR 2027}
   \def\And{\end{tabular}\hfil\linebreak[0]\hfil
            \begin{tabular}[t]{l}\bf\rule{\z@}{24pt}\ignorespaces}%
  \def\AND{\end{tabular}\hfil\linebreak[4]\hfil
            \begin{tabular}[t]{l}\bf\rule{\z@}{24pt}\ignorespaces}%
    \begin{tabular}[t]{l}\bf\rule{\z@}{24pt}Anonymous authors\\Paper under double-blind review\end{tabular}%
\fi
\vskip 0.3in minus 0.1in}}
\makeatother

\begin{document}
\raggedbottom

\maketitle

\AddToShipoutPictureFG*{%
  \AtTextUpperLeft{%
    \raisebox{24pt}[0pt][0pt]{%
      \makebox[\textwidth][r]{%
        \includegraphics[height=22pt]{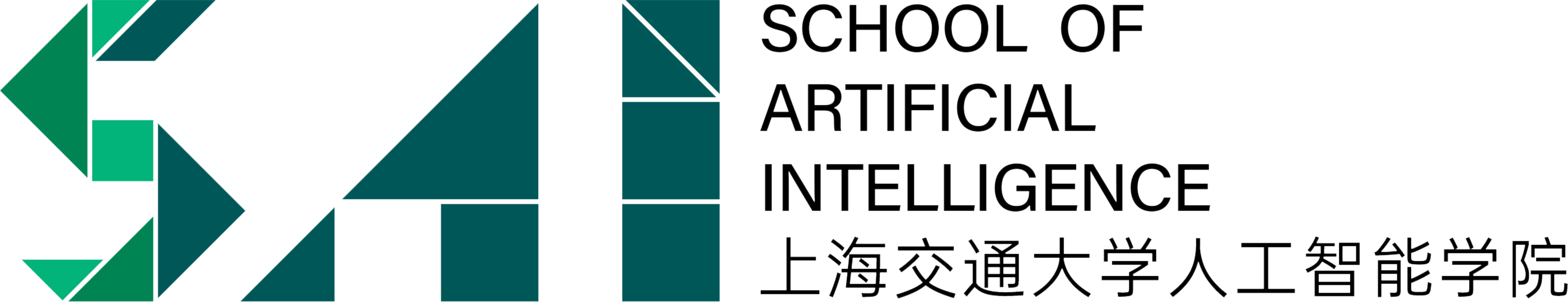}%
        \hspace{12pt}%
        \includegraphics[height=22pt]{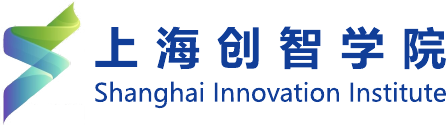}%
        \hspace{12pt}%
        \includegraphics[height=24pt]{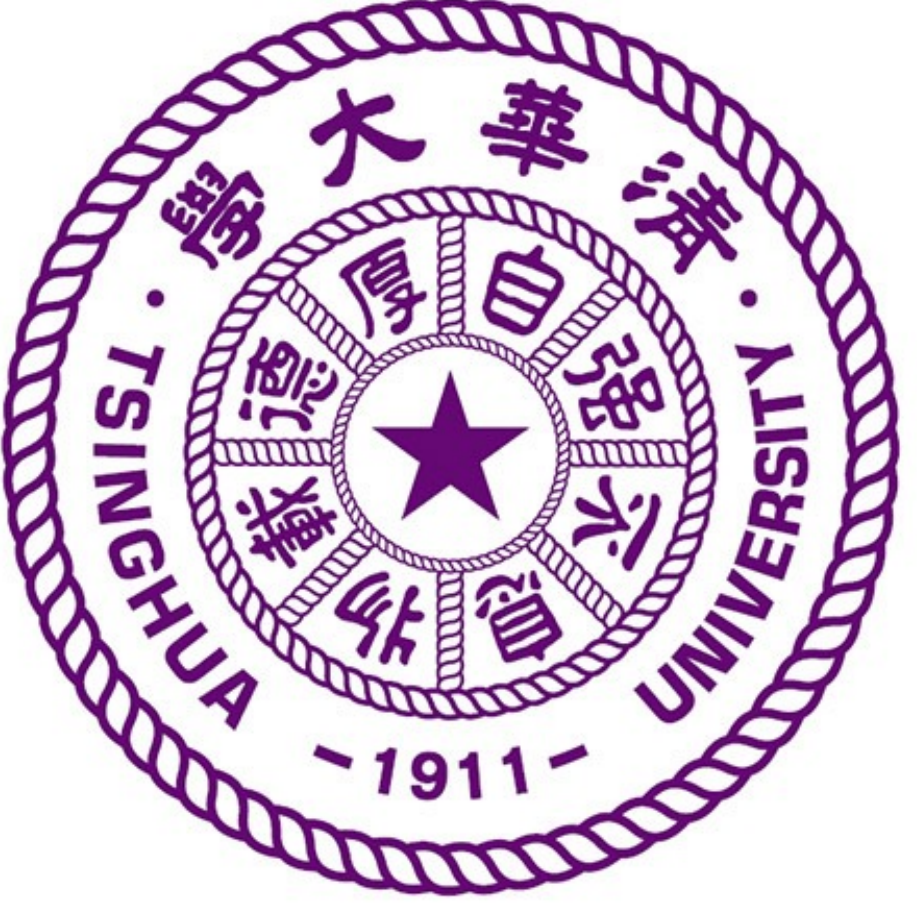}%
      }%
    }%
  }%
}

\begin{abstract}
Computer use agents are expanding from single-device operation toward cross-device systems that coordinate tasks across heterogeneous environments. Execution conditions are often only partially known at planning time and revealed through interaction. Failures may require intra-device modality switching or inter-device reassignment; failing to distinguish these cases can lead to repeated failures or premature termination. However, existing systems primarily scale up single-device agents without sufficiently distinguishing device-level and modality-specific execution conditions. We propose OctoNest, which coordinates stateful cross-device orchestration and iterative device-local modality control. Device Agents refine subtasks and select modalities, while an Orchestrator uses execution feedback to revise plans and device assignments. We also introduce CAPEBench, comprising 158 instances from 23 cross-device seed tasks with controlled perturbations. OctoNest leads all three quality metrics, improving Perfect Pass over the strongest baseline by 18.35 percentage points and reducing token cost per perfect pass by 39.8\%. Further analyses support the complementary roles of local refinement and global revision and demonstrate CAPEBench's ability to distinguish control limitations under changing execution conditions.
\end{abstract}

\section{Introduction}
\begin{figure}[htbp]
    \centering
    \includegraphics[width=0.9\linewidth]{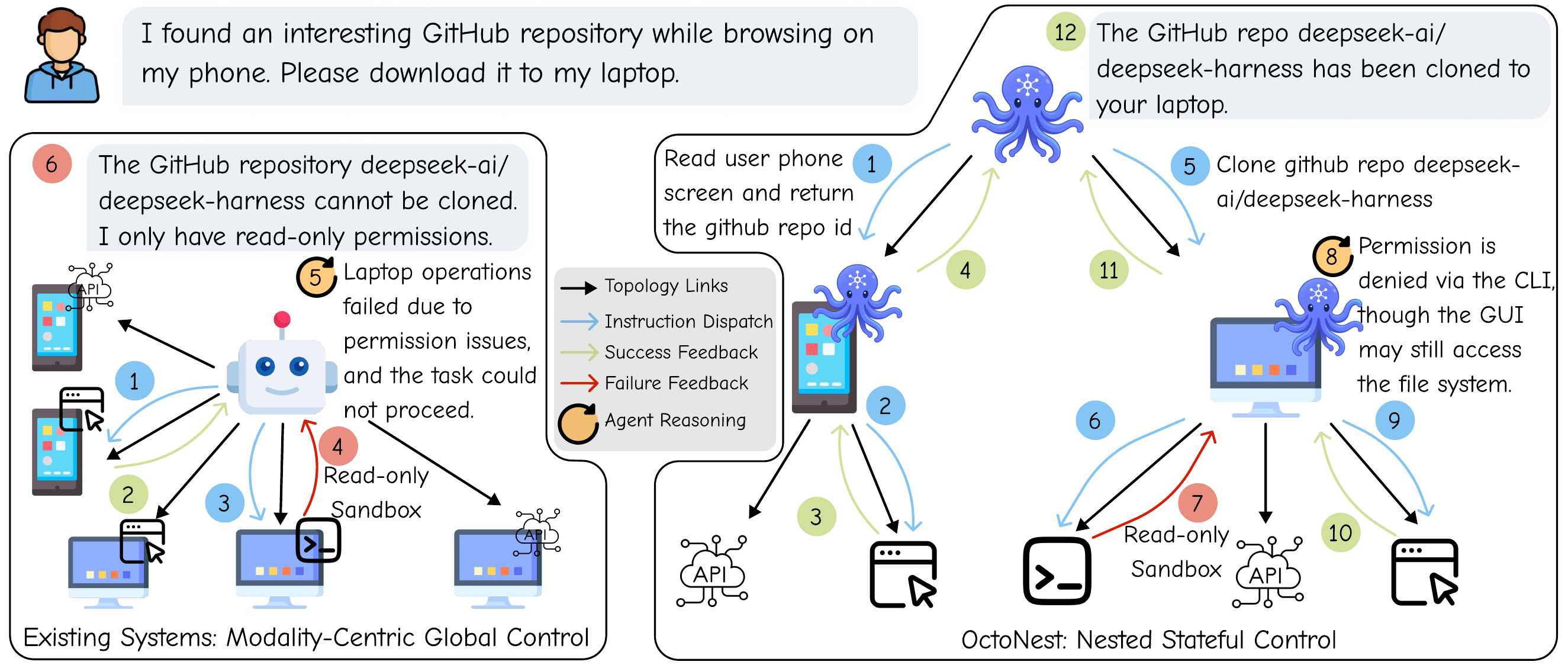}
    \caption{Modality-centric global control versus OctoNest's stateful cross-device execution, which distinguishes modality feasibility from device feasibility to guide intra-device and inter-device switching.}
    \label{fig:teaser}
\end{figure}

Recent advances in LLMs have extended them beyond conversational response generation toward agents that perceive and act in external environments~\citep{yao2023react, toolformer}. Among them, computer use agents, which operate digital applications and systems on users' behalf, have become a rapidly growing research direction~\citep{webarena, androidworld, theagentcompany}. They act through three primary execution modalities: 1) GUIs across web, desktop, and mobile~\citep{uitars2, mobileagentv35, androidworld}; 2) structured APIs for direct application and service operations~\citep{appworld, mcpworld}; and 3) CLIs for code, shell commands, and system operations~\citep{sweagent, terminalbench}. Recent systems further combine multiple execution modalities, including hybrid API--GUI control and GUI interaction augmented with Python or Bash execution~\citep{mcpworld, ufo2, coact}. Within a single device, these modalities provide complementary ways to accomplish a task.

In practice, many user requests span multiple devices because task-relevant information, applications, and execution capabilities are distributed across phones, personal computers, and remote machines~\citep{crab,jarvisgui2026,devicesworld2026}. CRAB centralizes perception across environments, planning, and action selection in a primary agent~\citep{crab}. TopoClaw similarly relies on a primary agent to coordinate multiple devices while making both planning and execution decisions~\citep{topoclaw}. UFO\textsuperscript{3} separates global graph orchestration from device execution, maintaining a mutable task graph and dispatching its nodes to device agents tied to device-specific execution modalities~\citep{ufo3}. Despite their different organizations, these systems coordinate endpoints bound to a modality, grounding cross-device decisions into the action spaces exposed by their modalities and returning observations to the global agent.

Cross-device execution depends on both global task dependencies and execution states local to each device and its modalities. However, existing systems largely leave the global agent to interpret modality-level observations for cross-device coordination, without sufficiently distinguishing between adaptation within a device and reassignment across devices. For example, a write operation blocked by a CLI's read-only sandbox may still be completed through the GUI on the same device, whereas a device-wide network disconnection may make a network-dependent subtask infeasible on that device. Access to modalities subject to different runtime constraints can enable continued execution when one modality becomes unavailable~\citep{toolcua2026,guivscli2026}, while failures that leave no feasible alternative on the current device require reassignment. The states governing these possibilities are only partially known at planning time and progressively revealed through interaction~\citep{toolmaze,windowsagentarena}, making execution a failure-prone process of state discovery. This motivates supporting multiple modalities and updating execution state from feedback to distinguish modality feasibility from device feasibility, guiding intra-device or inter-device switching while preserving task constraints.

We propose \textbf{OctoNest}, a stateful cross-device execution framework that coordinates intra-device and inter-device switching through nested control. The orchestrator maintains orchestration state to plan and assign device-level subtasks. Each device agent maintains device-local state to progressively refine its subtask, select API, CLI, or GUI modality executors, and switch modalities when needed. When no viable local continuation can be identified, it returns execution evidence to the orchestrator, which revises the remaining plan and device assignments. As illustrated in Figure~\ref{fig:teaser}, this design preserves progress through local adaptation when possible and enables reassignment when necessary, supporting continued execution under changing conditions.

We introduce \textbf{CAPEBench} (\emph{Cross-device Adaptation to Perturbations in Execution}), comprising 158 instances derived from 23 seed tasks. CAPEBench covers unperturbed, modality-level, device-level, and mixed conditions, with modality-level perturbations testing intra-device switching, device-level perturbations testing inter-device switching, and mixed perturbations testing both. Compared with the strongest baseline, OctoNest improves Completion, Adherence, and Perfect Pass by 10.51, 11.36, and 18.35 percentage points, respectively, while reducing token cost per perfect pass by 39.8\%. Further analysis shows that OctoNest selects capability-compatible modality sequences and achieves higher affected-step recovery under perturbations, supporting the complementary roles of device-local control and cross-device orchestration.

Our contributions are threefold:
\begin{itemize}[leftmargin=*]
\item We propose \textbf{OctoNest}, a stateful cross-device execution framework that unifies intra-device and inter-device switching. By distinguishing modality feasibility from device feasibility, OctoNest enables agents to adapt at the appropriate level as state is revealed during execution.

\item We distinguish device-local modality decisions from cross-device planning: device agents use execution feedback to refine subtasks, select and combine API, CLI, and GUI modalities, and perform intra-device switching, while the orchestrator uses device-level outcomes to revise the remaining plan and perform inter-device switching. This enables local recovery when an alternative modality remains and cross-device reassignment when local execution is no longer viable.

\item We introduce \textbf{CAPEBench}, a cross-device execution benchmark with controlled perturbations that diagnoses whether agents recognize interaction-revealed changes in device and modality state and choose appropriately between intra-device and inter-device switching. On CAPEBench, OctoNest improves Perfect Pass by 18.35 percentage points and cuts the token cost per perfect pass by 39.8\% over the strongest baseline.
\end{itemize}

\section{Methodology}
\label{Methodology}

\subsection{Problem Formulation}

We consider a natural-language instruction \(I\) to be completed over a set of heterogeneous devices \(\mathcal{D}\), subject to task dependencies and device assignment constraints \(\mathcal{C}_I\). Each device \(d \in \mathcal{D}\) exposes a set of execution modalities \(\mathcal{M}_d\), such as API, CLI, and GUI. At time \(t\), its latent execution state is represented as
\begin{equation*}
x_d^t =
\left(
x_{d,\mathrm{dev}}^t,
\{x_{d,m}^t\}_{m\in\mathcal{M}_d}
\right)
\end{equation*}
where \(x_{d,\mathrm{dev}}^t\) captures conditions shared across modalities on device \(d\), such as connectivity, while \(x_{d,m}^t\) captures modality-specific conditions, such as access permissions and authentication.

For an unfinished subtask goal \(g\), let \(\mathcal{R}_d(g;x_d^t)\) denote the set of feasible executions that complete \(g\) on device \(d\) under the current state and applicable task constraints. Each execution may involve one or more modalities.

Within a device, the failure of one modality does not imply that \(\mathcal{R}_d(g;x_d^t)\) is empty. If \(g\) remains feasible using an alternative modality, execution can continue through \emph{intra-device switching}, which changes modalities while retaining the subtask's device assignment. Across devices, when
\begin{equation*}
\mathcal{R}_d(g;x_d^t)=\varnothing,
\qquad
\exists d'\in\mathcal{D}\setminus\{d\}:
\mathcal{R}_{d'}(g;x_{d'}^t)\neq\varnothing
\end{equation*}
completing the subtask requires \emph{inter-device switching}, which reassigns the unfinished goal from \(d\) to a feasible device \(d'\). Both forms of switching must respect \(\mathcal{C}_I\).

These feasibility conditions depend on latent execution states that are only partially known at planning time~\citep{bench2robust2026}. Agents must therefore use evidence revealed through interaction to assess whether execution can continue within the current device or requires reassignment across devices. The objective is to complete \(I\) while satisfying \(\mathcal{C}_I\), adapting execution through intra-device or inter-device switching as task-relevant state is revealed.

\subsection{OctoNest Overview}

\begin{figure}[t]
    \centering
    \includegraphics[width=0.9\linewidth]{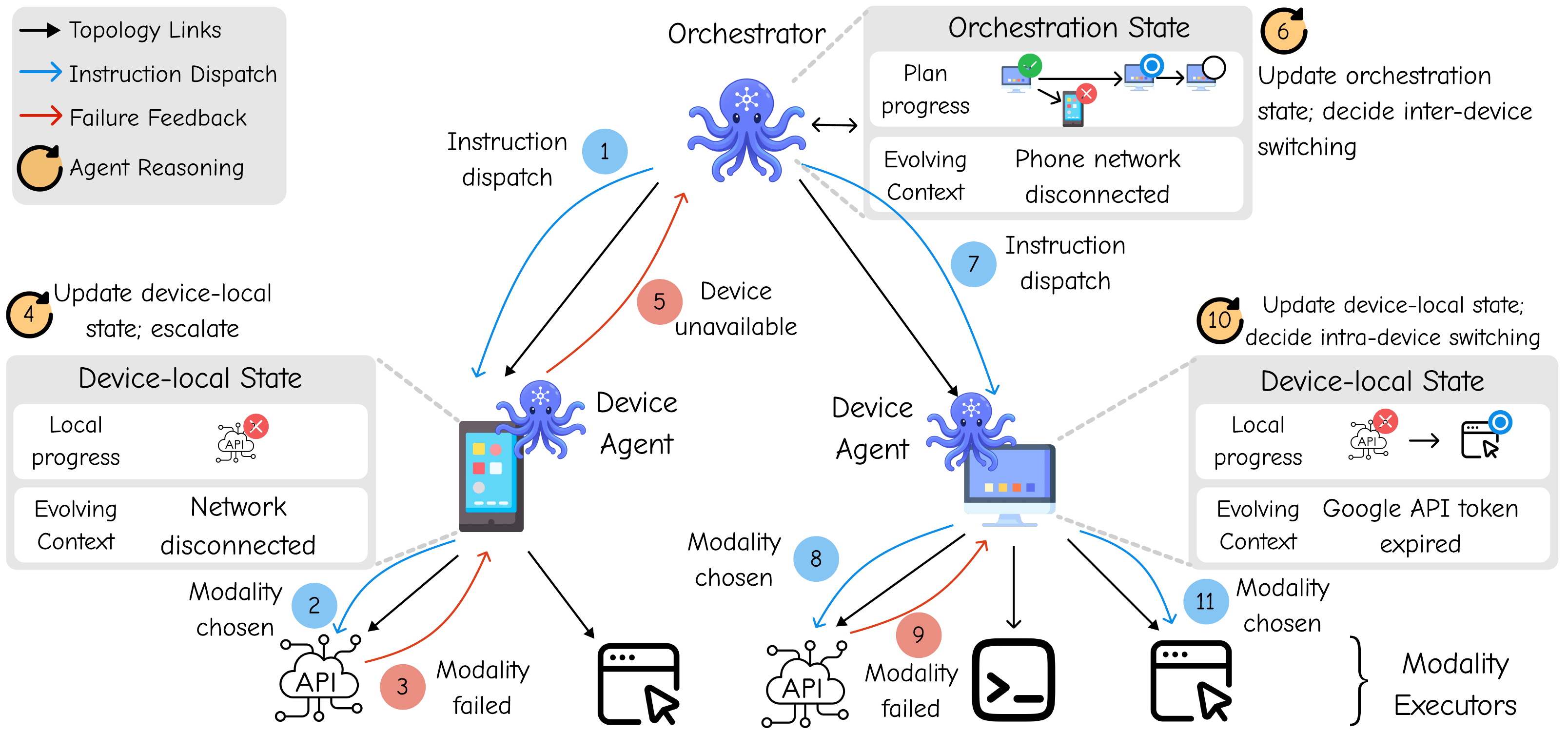}
    \caption{Overview of OctoNest. The Orchestrator maintains orchestration state for device-level coordination and inter-device switching, while each Device Agent maintains device-local state to progressively ground subtasks and coordinate API, CLI, and GUI Modality Executors.}
    \label{fig:octonest-overview}
\end{figure}

As shown in Figure~\ref{fig:octonest-overview}, OctoNest maintains execution state at two levels: orchestration state and device-local state. The Orchestrator uses orchestration state to plan and assign device-level subtasks. Each Device Agent uses device-local state to refine its subtask into instructions for API, CLI, or GUI Modality Executors, incorporating their feedback to guide execution. When feedback indicates that another modality offers a viable continuation, it performs intra-device switching; when no viable local continuation can be identified, it returns execution evidence to the Orchestrator. Device-level outcomes update orchestration state, enabling the Orchestrator to propagate results, revise the remaining plan, and perform inter-device switching when reassignment is needed.

\subsection{Orchestration State and Inter-Device Switching}

Cross-device coordination requires tracking subtask dependencies, propagating intermediate results, and accounting for device feasibility~\citep{ufo3}, while revising the remaining plan as execution reveals new evidence~\citep{adaplanner}. The Orchestrator supports these requirements through orchestration state, represented at coordination step \(t\) as
\begin{equation*}
s_t^O =
\left(
P_t,
\mathcal{Y}_t,
\mathcal{F}_t
\right)
\end{equation*}
where \(P_t\) is the current global plan, \(\mathcal{Y}_t\) contains completed subtask results, and \(\mathcal{F}_t\) contains execution feedback and context available for assessing device feasibility and revising the remaining plan. The plan is represented as a dependency-ordered subtask chain
\begin{equation*}
P_t = \langle q_1,\ldots,q_K\rangle,
\qquad
q_j = (g_j,d_j,c_j,z_j)
\end{equation*}
where \(K\) is the number of subtasks in the current plan, \(g_j\) specifies the subtask objective, \(d_j\) is its assigned device, \(c_j\) contains execution context required by the subtask, and \(z_j\) denotes its execution status.

Given the instruction, task constraints, and available device profiles, the Orchestrator initializes \(P_0\) and dispatches the next executable subtask to its assigned Device Agent. After local execution terminates, the Device Agent returns an outcome \(r_j\), containing the execution status, available results, and execution context. This outcome updates the orchestration state:
\begin{equation*}
s_{t+1}^O =
\mathcal{U}_O
\left(
s_t^O,
q_j,
r_j
\right)
\end{equation*}
where \(\mathcal{U}_O\) denotes the orchestration state update. For a successful outcome, the Orchestrator records the result and propagates required information to dependent subtasks. On escalation, it uses the returned evidence to revise the unfinished plan, potentially refining subtasks or reassigning unfinished work to another device consistent. Reassigning unfinished work to a different device constitutes inter-device switching.

\subsection{Device Agent: Device-Local State and Intra-Device Switching}

Assigning a subtask to a device leaves open how it should be carried out through the available modalities. The next executable step depends on local progress and modality conditions revealed through interaction, so execution requires progressively refining the subtask and adapting modality choices as evidence accumulates~\citep{adapt,ufo2}. The Device Agent manages this process by maintaining device-local state and coordinating the available Modality Executors. For a subtask \(q_j\) assigned to device \(d=d_j\), its state \(s^d_{j,\ell}\) at local step \(\ell\) retains prior execution attempts, their outcomes, and the interaction context used to assess local progress and modality feasibility.

At each local step, the Device Agent makes one of three decisions:
\begin{equation*}
\pi_d\left(q_j,s^d_{j,\ell},\mathcal{M}_d\right)
\rightarrow
\big\{\textsc{Execute}(m,u),
      \textsc{Done}(y_j),
      \textsc{Escalate}(e_j)\big\}
\end{equation*}
where \(\pi_d\) denotes the decision process of the Device Agent on device \(d\). \textsc{Execute} issues instruction \(u\) through modality \(m\), \textsc{Done} returns the completed subtask result \(y_j\), and \textsc{Escalate} returns execution evidence \(e_j\) for orchestration-level replanning.

At local step \(\ell\), an \textsc{Execute} decision selects modality \(m\in\mathcal{M}_d\) and issues instruction \(u\). The selected Modality Executor returns execution feedback \(\mathcal{E}_{d,m}(u)\), which updates the device-local~state:
\begin{equation*}
s^d_{j,\ell+1}
=
\mathcal{U}_D\big(s^d_{j,\ell},m,u,\mathcal{E}_{d,m}(u)\big),
\end{equation*}
where \(\mathcal{U}_D\) denotes the device-local state update.

A successful modality step does not necessarily complete the device-level subtask; the Device Agent checks remaining requirements and generates another step when needed. After a failed step, the Device Agent uses execution feedback to revise the instruction or select an alternative modality; the latter constitutes intra-device switching. If no viable local continuation is identified or the local execution budget is exhausted, it escalates to the Orchestrator. A \textsc{Done} or \textsc{Escalate} decision terminates local execution and produces the device-level outcome \(r_j\), including execution context.

\subsection{Modality Executors}

API, CLI, and GUI execution involve distinct action spaces and observations~\citep{appworld,androidworld}. OctoNest uses a dedicated Modality Executor for each modality to translate instructions into concrete actions and interpret their feedback~\citep{planact,yao2023react}, separating modality-specific execution from device-local coordination. Given an instruction \(u\), the executor for modality \(m\) on device \(d\) performs one or more actions and returns
\begin{equation*}
\mathcal{E}_{d,m}(u) \rightarrow (\sigma,y,o),
\end{equation*}
where \(\sigma\) indicates execution success or failure, \(y\) contains any available output, and \(o\) records execution observations, including failure details when applicable. This feedback is incorporated into device-local state to guide subsequent execution. The API Executor invokes structured application or service operations; the CLI Executor performs command-line, computation, and file-system operations; and the GUI Executor interacts with applications through visual observations and interface actions.

\section{Experiments}
\label{experiment}

\begin{figure}[t]
    \centering
    \includegraphics[width=0.9\linewidth]{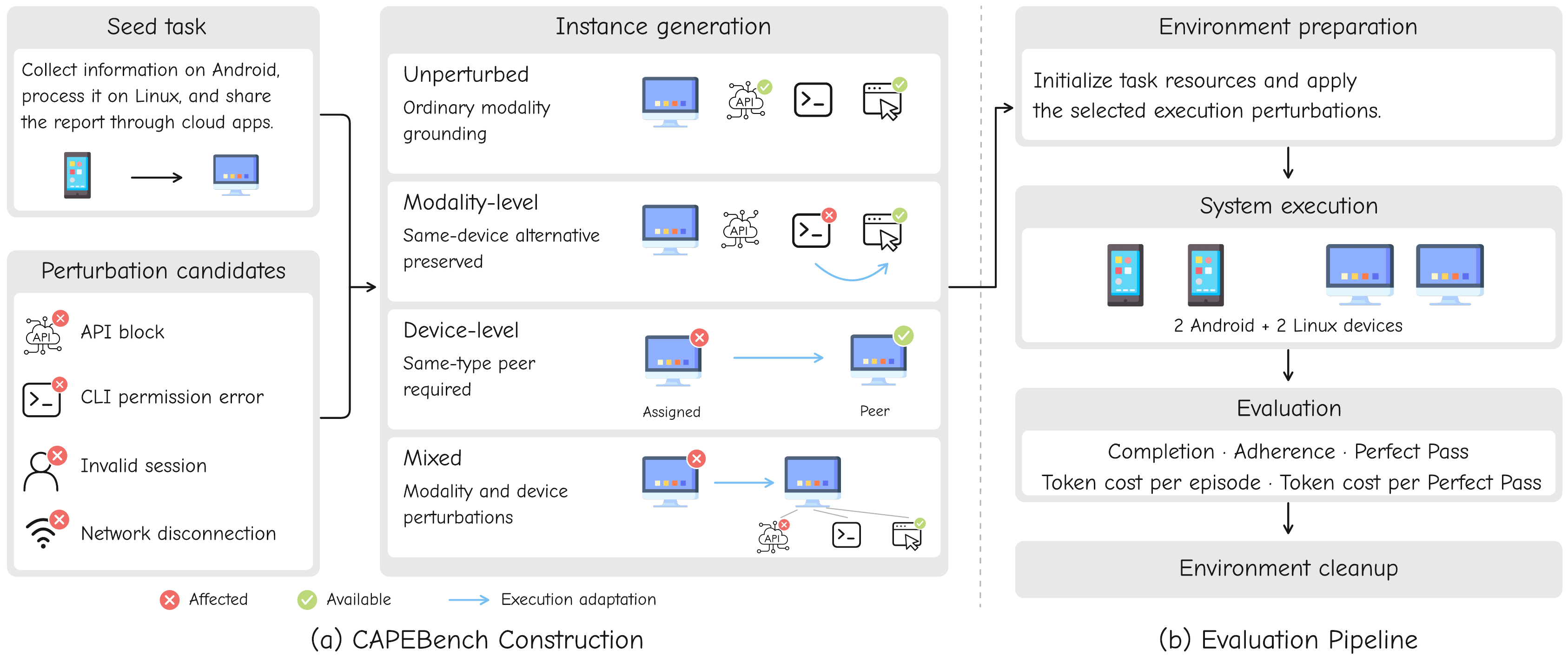}
    \caption{CAPEBench construction (left) and evaluation pipeline (right).}
    \label{fig:benchmark}
\end{figure}

\subsection{CAPEBench and Experimental Setup}
\label{sec:benchmark}

CAPEBench comprises 158 instances derived from 23 cross-device seed tasks, each instantiated under a specified runtime condition. It runs on two Linux and two Android devices with real application services and local file systems. Each device exposes a subset of API, CLI, and GUI modalities, with only partial service API coverage, requiring agents to ground task steps through the available modalities. As illustrated in Figure~\ref{fig:benchmark}, CAPEBench covers unperturbed, modality-level, device-level, and mixed conditions~\citep{toolmaze,failtalms,masfire}. Unperturbed instances evaluate ordinary modality grounding. Modality-level perturbations preserve a same-device alternative to test intra-device switching; device-level perturbations prevent local completion while preserving a same-type peer to test inter-device switching; mixed perturbations test both. Perturbations are applied deterministically to the environment, with task-specific effects revealed through execution.

Following prior agent evaluations~\citep{agentboard}, we report three quality metrics. \textbf{Task Completion} (Comp.) measures the percentage of required service and file-state postconditions satisfied. \textbf{Instruction Adherence} (Adh.) measures the fraction of gold steps completed with the intended semantics on an allowed device; modality-level perturbations retain the original device constraint, whereas device-level perturbations permit a same-type peer. \textbf{Perfect Pass} (PP) measures the percentage of episodes in which every gold step is completed with the intended semantics on an allowed device. Efficiency metrics are average tokens per episode (\textbf{Tok./Ep.}) and tokens per perfect pass (\textbf{Tok./PP}), computed as Tok./Ep. divided by the perfect-pass rate.\footnote{Tok./PP includes tokens spent on unsuccessful episodes and is reported as $\infty$ when no episode achieves Perfect Pass.}

We compare three representative cross-device systems. CRAB pioneers coordinated interaction across desktop and mobile environments~\citep{crab}. UFO\textsuperscript{3} supports adaptive execution through explicit task dependencies and feedback-driven task-graph updates~\citep{ufo3}; we extend it with API, CLI, and GUI execution to form UFO\textsuperscript{3}-Hybrid. TopoClaw supports topology-aware coordination through a primary agent that accounts for device capabilities and user context~\citep{topoclaw}. All methods use a 30-minute execution timeout. Unless otherwise stated, text-based agents use DeepSeek-V4-Pro and vision-based agents use Kimi K2.5. Main comparisons use \textbf{Full-158}; selected analyses use a fixed sample of 82 instances.

\subsection{RQ1: How effective and efficient is OctoNest?}
\label{sec:main-results}

\begin{table}[t]
\centering
\caption{Main results on CAPEBench. Bold and underlined values mark the best and second-best results. Brackets show pointwise 95\% confidence intervals for changes from paired seed-task bootstrap.}
\label{tab:main-results}
\small
\setlength{\tabcolsep}{4pt}
\renewcommand{\arraystretch}{1.16}
\begin{tabularx}{\linewidth}{@{}
l
>{\raggedright\arraybackslash}X
*{5}{>{\raggedleft\arraybackslash}X}
@{}}
\toprule
\multicolumn{2}{c}{\textbf{System}} &
\multicolumn{3}{c}{\textbf{Quality}} &
\multicolumn{2}{c}{\textbf{Efficiency}} \\
\cmidrule(lr){1-2}\cmidrule(lr){3-5}\cmidrule(l){6-7}
\textbf{Method} & \textbf{Exec.} &
\textbf{Comp. (\%)} $\uparrow$ &
\textbf{Adh. (\%)} $\uparrow$ &
\textbf{PP (\%)} $\uparrow$ &
\textbf{Tok./Ep.} $\downarrow$ &
\textbf{Tok./PP} $\downarrow$ \\
\midrule
CRAB & GUI & 2.37 & 16.19 & 0.00 & 545,102 & $\infty$ \\
CRAB & API & 22.95 & 38.39 & 0.00 & \underline{513,463} & $\infty$ \\
\midrule
TopoClaw & Hybrid & 46.86 & 52.17 & 5.70 & 933,484 & 16,387,826 \\
\midrule
UFO\textsuperscript{3} & GUI & 47.06 & 50.94 & 10.13 & 1,452,661 & 14,345,029 \\
UFO\textsuperscript{3} & API & 59.28 & 65.44 & 0.00 & \textbf{426,608} & $\infty$ \\
UFO\textsuperscript{3} & Hybrid & \underline{65.00} & \underline{65.46} & \underline{22.78} & 702,893 & \underline{3,084,917} \\
\midrule
\rowcolor[HTML]{F0EDF9}
\textbf{OctoNest} & Hybrid & \textbf{75.51} & \textbf{76.82} & \textbf{41.14} & 764,552 & \textbf{1,858,449} \\
\midrule
\multicolumn{2}{l}{\textit{Change vs.\ UFO\textsuperscript{3}-Hybrid}} & \textcolor{gainGreen}{$+10.51$} & \textcolor{gainGreen}{$+11.36$} & \textcolor{gainGreen}{$+18.35$} & \textcolor{lossRed}{$+8.8\%$} & \textcolor{gainGreen}{$-39.8\%$} \\
\multicolumn{2}{l}{\scriptsize 95\% CI of changes} &
{\scriptsize $[2.79,\,18.22]$} &
{\scriptsize $[5.03,\,17.69]$} &
{\scriptsize $[7.33,\,28.75]$} &
{\scriptsize $[-9.4,31.9]\%$} &
{\fontsize{6.5}{7}\selectfont $[-58.8,-11.6]\%$} \\
\bottomrule
\end{tabularx}
\end{table}

\begin{table}[t]
\centering
\footnotesize
\setlength{\tabcolsep}{4.2pt}
\renewcommand{\arraystretch}{1.12}
\caption{Quality and efficiency across runtime conditions compared with UFO\textsuperscript{3}-Hybrid.}
\label{tab:condation-comparison}
\begin{tabular*}{\textwidth}{
@{\extracolsep{\fill}}llrrrr@{}
}
\toprule
\textbf{Condition} &
\textbf{Method} &
\textbf{Comp. (\%)} $\uparrow$ &
\textbf{Adh. (\%)} $\uparrow$ &
\textbf{PP (\%)} $\uparrow$ &
\textbf{Tok./PP} $\downarrow$ \\
\midrule
\multirow{2}{*}{Unperturbed}
& UFO\textsuperscript{3}-Hybrid
& 73.19
& 70.97
& 30.43
& 2,146,184 \\
& OctoNest
& \textbf{81.81}\increaseGain{11.8}
& \textbf{86.05}\increaseGain{21.2}
& \textbf{52.17}\increaseGain{71.4}
& \textbf{1,151,222}\decreaseGain{46.4} \\
\midrule
\multirow{2}{*}{Modality}
& UFO\textsuperscript{3}-Hybrid
& 82.98
& 79.43
& 38.10
& 1,775,644 \\
& OctoNest
& \textbf{84.64}\increaseGain{2.0}
& \textbf{85.94}\increaseGain{8.2}
& \textbf{54.76}\increaseGain{43.8}
& \textbf{1,342,100}\decreaseGain{24.4} \\
\midrule
\multirow{2}{*}{Device}
& UFO\textsuperscript{3}-Hybrid
& 55.69
& 60.79
& 19.44
& 3,321,659 \\
& OctoNest
& \textbf{62.59}\increaseGain{12.4}
& \textbf{68.41}\increaseGain{12.5}
& \textbf{33.33}\increaseGain{71.4}
& \textbf{2,644,824}\decreaseGain{20.4} \\
\midrule
\multirow{2}{*}{Mixed}
& UFO\textsuperscript{3}-Hybrid
& 54.33
& 55.89
& 10.53
& 7,395,304 \\
& OctoNest
& \textbf{74.39}\increaseGain{36.9}
& \textbf{71.68}\increaseGain{28.3}
& \textbf{31.58}\increaseGain{200.0}
& \textbf{2,465,462}\decreaseGain{66.7} \\
\bottomrule
\end{tabular*}
\end{table}

Table~\ref{tab:main-results} shows that OctoNest achieves the highest Completion, Adherence, and Perfect Pass among the evaluated systems. Compared with UFO\textsuperscript{3}-Hybrid, it improves all three quality metrics, including a gain of 18.35 percentage points in Perfect Pass, with the pointwise 95\% confidence interval for each improvement excluding zero. Although mean token usage per episode is higher, token cost per perfect pass decreases by 39.8\%. The improvement in efficiency therefore comes from a higher yield of successful executions, rather than lower token expenditure on each attempt. The results suggest that execution organization matters alongside modality availability. CRAB and TopoClaw rely on primary-agent control and achieve limited end-to-end success, even with TopoClaw's hybrid execution. UFO\textsuperscript{3}-Hybrid separates cross-device planning from device-level execution and achieves substantially higher Perfect Pass, consistent with the value of hierarchical task organization. OctoNest further improves Perfect Pass despite sharing API, CLI, and GUI access with UFO\textsuperscript{3}-Hybrid. This gap motivates examining how device-local modality decisions interact with cross-device orchestration, using UFO\textsuperscript{3}-Hybrid as the primary behavioral comparison.

Table~\ref{tab:condation-comparison} shows improvements in all three quality metrics and lower token cost per perfect pass across all four runtime conditions. Without perturbations, OctoNest achieves higher Perfect Pass, indicating that the benefits extend to ordinary task execution, where agents must still select suitable modalities and satisfy device constraints. Under modality-level perturbations, similar Completion but substantially different Perfect Pass reveals a gap between average final state satisfaction and fully adherent execution. OctoNest's advantage in this condition is particularly apparent in completing the entire required execution correctly, where intra-device switching can preserve progress after a modality becomes unavailable. Device-level and mixed perturbations additionally require agents to continue execution beyond the originally assigned device. OctoNest achieves higher Perfect Pass under device-level perturbations. Under mixed perturbations, which exercise both intra-device and inter-device switching, it improves both Completion and Perfect Pass. Higher success also reduces token cost per perfect pass by 66.7\%, the largest reduction among the four conditions. Nevertheless, Perfect Pass remains below one-third in this setting, showing that reliable execution under combined modality and device constraints remains challenging.

\subsection{RQ2: Which components contribute to OctoNest's performance?}
\label{sec:nested-control-analysis}

\begin{wraptable}{r}{0.50\textwidth}
\vspace{-\baselineskip}
\centering
\caption{Architectural ablations. Quality metrics are percentages; Tok./PP is in millions.}
\label{tab:ablation-results}
\scriptsize
\setlength{\tabcolsep}{3pt}
\resizebox{\linewidth}{!}{%
\begin{tabular}{lrrrr}
\toprule
Method & Comp. & Adh. & PP & Tok./PP \\
\midrule
OctoNest & \textbf{75.55} & \textbf{77.49} & \textbf{43.90} & \textbf{1.631} \\
\midrule
No Device Agent & 56.59 & 58.02 & 18.29 & 2.237 \\
One-shot Device Agent & 57.52 & 56.21 & 14.63 & 3.778 \\
Unified Device Agent & 51.30 & 55.51 & 18.29 & 3.599 \\
Reactive Orchestrator & 62.70 & 64.47 & 32.93 & 1.931 \\
Fixed orchestration plan & 40.14 & 47.68 & 21.95 & 2.214 \\
Failure-only device feedback & 46.20 & 50.38 & 30.49 & 2.012 \\
\bottomrule
\end{tabular}%
}
\end{wraptable}

We first examine the contributions of device-local and orchestration-level control on the 82-instance sample (Table~\ref{tab:ablation-results}). Removing the Device Agent substantially reduces Perfect Pass, showing the value of refining assigned subtasks before and during modality-specific execution. Without this layer, the Orchestrator must manage cross-device task progress while also issuing instructions directly to Modality Executors. A one-shot Device Agent retains the intermediate refinement step but reduces Perfect Pass from 43.90\% to 14.63\%: its initial modality choice and execution instruction cannot be revised in response to subsequent observations. Unifying execution across all modalities within a single Device Agent also substantially reduces Perfect Pass, supporting the separation of device-local coordination from modality-specific execution. Together, they support iterative device-local control that uses feedback to refine instructions and modality choices while delegating concrete actions to modality-specific executors.

At the orchestration level, the Reactive Orchestrator retains execution history but chooses successive assignments without maintaining an explicit plan for the remaining task. Its lower performance supports the value of organizing future assignments and their dependencies, providing a persistent structure for coordinating progress across devices. The fixed-plan variant preserves this structure but cannot revise it when execution reveals that an assignment is no longer viable. Its substantial degradation shows why device-local adaptation alone is insufficient: when local execution escalates a failure, the fixed plan cannot introduce recovery assignments, and execution terminates. These results support an Orchestrator that combines an explicit future plan with execution-conditioned revision. Failure-only device feedback further shows that revision requires evidence about what went wrong. Audited traces reveal an approximately even split between transferring failed subtasks and retrying them on the original device, reflecting difficulty distinguishing device infeasibility from locally recoverable execution problems. Repeated attempts add overhead, and many tasks are aborted after a second failure. Informative device feedback therefore helps the Orchestrator determine when to preserve an assignment and when inter-device switching is needed.

\subsection{RQ3: How does OctoNest adapt during execution?}

\begin{wrapfigure}[13]{r}{0.4\columnwidth}
  \centering
  \setlength{\abovecaptionskip}{4pt}
  \setlength{\belowcaptionskip}{0pt}
  \includegraphics[
    width=\linewidth,
    trim=0 14bp 0 0,
    clip
  ]{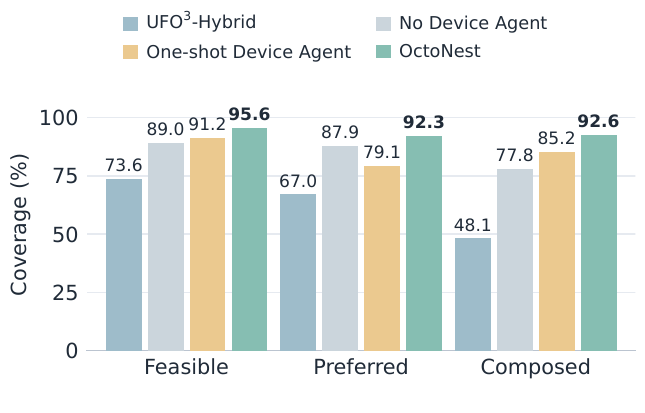}
  \caption{Modality-sequence coverage on unperturbed single-device task parts.}
  \label{fig:modality-grounding}
\end{wrapfigure}

\paragraph{Progressive modality grounding.}
We examine how agents select and combine modalities to complete the parts of unperturbed CAPEBench instances that require execution on a single device. \emph{Feasible} measures coverage of any sequence capable of satisfying the task requirements. \emph{Preferred} measures coverage of the annotated sequence expected to fulfill these requirements straightforwardly and efficiently. \emph{Composed} restricts preferred coverage to task parts requiring multiple modalities in the annotated preferred sequence. Coverage requires the reference modalities to appear in order within the observed execution sequence; unreached parts count as uncovered.

Figure~\ref{fig:modality-grounding} shows higher feasible and composed coverage for OctoNest than UFO\textsuperscript{3}-Hybrid, indicating better selection of capability-compatible sequences and composition of complementary modalities. Higher preferred coverage than the one-shot Device Agent supports progressively decomposing subtasks into finer steps with suitable modalities, rather than requiring one modality to complete the entire subtask. These gains arise without injected failures, showing that device-local refinement benefits ordinary execution as well as recovery.

\begin{wrapfigure}{r}{0.45\columnwidth}
  \centering
  \setlength{\abovecaptionskip}{4pt}
  \setlength{\belowcaptionskip}{0pt}
  \includegraphics[
  width=\linewidth,
  trim=0 6bp 0 0,
  clip
]{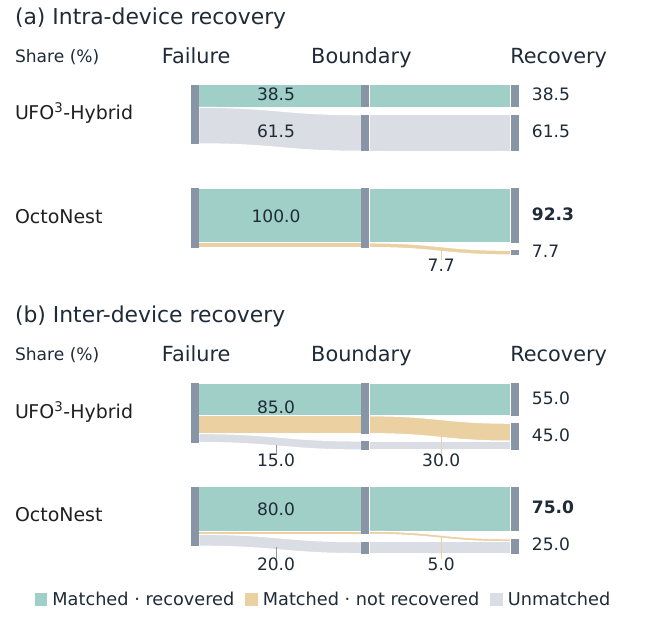}
  \caption{Intra-device (a) and inter-device (b) recovery on instances where both systems encounter the induced failure.}
  \label{fig:recovery-flow}
\end{wrapfigure}

\paragraph{Intra-device and inter-device recovery.} We evaluate intra-device and inter-device recovery under single modality-level and device-level perturbations, respectively, restricting analysis to cases with a recovery destination where both systems encounter the failure. \emph{Boundary match} measures whether recovery stays on the assigned device under modality-level perturbations or moves to the required peer under device-level perturbations. \emph{Affected-step recovery} additionally requires every affected gold step to be completed with full adherence.

Under modality-level perturbations, OctoNest consistently keeps recovery on the assigned device and achieves higher affected-step recovery than UFO\textsuperscript{3}-Hybrid (Figure~\ref{fig:recovery-flow}(a)). This shows that local alternatives can support continued execution without unnecessary device reassignment. Under device-level perturbations, OctoNest achieves higher affected-step recovery despite a slightly lower boundary match rate (Figure~\ref{fig:recovery-flow}(b)). The orange branch shows that more baseline cases reach the required peer but still fail to recover, highlighting the importance of successful continuation after reassignment. In audited UFO\textsuperscript{3}-Hybrid traces, capability-mismatched modality choices and overly broad instructions prevent successful continuation, sometimes exhausting executor budgets. Corresponding OctoNest traces show the destination Device Agent breaking down the transferred subtask and selecting modalities suited to local execution conditions. Together with the orchestration ablations, these observations support complementary roles for the two control levels: the Orchestrator revises device assignments, while the Device Agent refines execution within the selected device.

\subsection{RQ4: How does CAPEBench diagnose control limitations?}
\label{sec:CAPEBench-validity}

We pair perturbation families with architectural interventions to examine whether CAPEBench distinguishes limitations in different control capabilities.

\begin{wrapfigure}{r}{0.50\textwidth}
  \centering
  \setlength{\abovecaptionskip}{4pt}
  \setlength{\belowcaptionskip}{0pt}
  \includegraphics[
  width=\linewidth,
  trim=0 6bp 0 0,
  clip
]{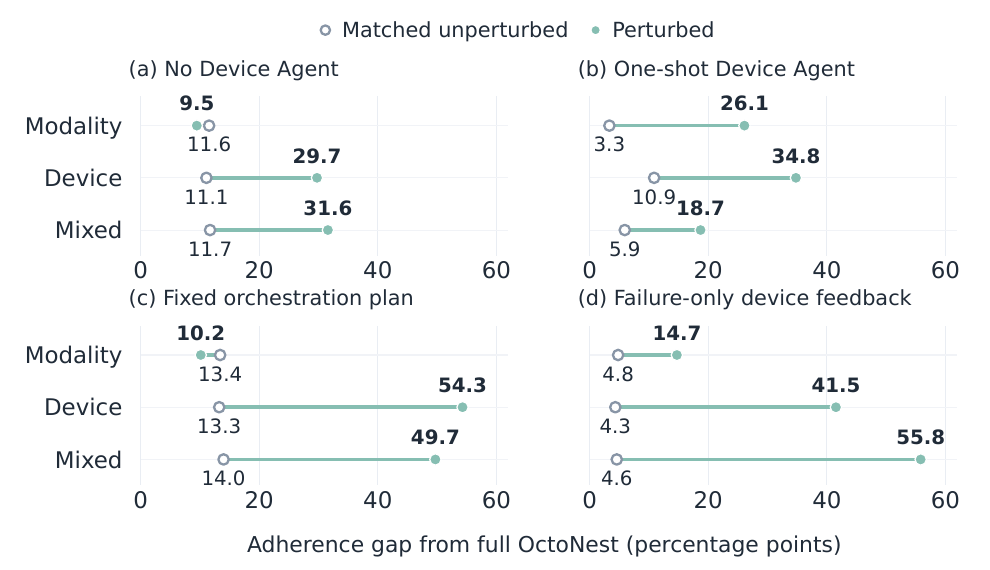}
  \caption{Adherence gaps from full OctoNest on matched unperturbed and perturbed tasks.}
  \label{fig:CAPEBench-diagnostics}
\end{wrapfigure}

\textbf{Diagnostic sensitivity.}
For each intervention, Figure~\ref{fig:CAPEBench-diagnostics} compares its Adherence gap from full OctoNest under perturbation with the corresponding gap on matched unperturbed tasks. The change between the two endpoints captures the additional penalty beyond the gap already present without perturbations. The interventions exhibit distinct responses. With a one-shot Device Agent, the gap widens substantially under modality-level perturbations, showing the importance of refining local execution when the initially selected modality becomes unavailable. With a fixed orchestration plan, the gap changes little under modality-level perturbations but widens by 41.0 and 35.7 percentage points under device-level and mixed perturbations, respectively. This contrast reflects the different control requirements: same-device alternatives can be handled locally, whereas device-level obstacles require a revised assignment. Failure-only device feedback exhibits its largest gap expansion under mixed perturbations, where deciding between local continuation and device reassignment requires informative execution evidence. Without such evidence, the Orchestrator has difficulty determining which control response is appropriate. Together, these patterns show that CAPEBench reveals how different control limitations respond to changing execution conditions, beyond the performance differences already visible on unperturbed tasks.

\section{Related Work}
\label{relwork}

LLM agents operate through heterogeneous execution modalities, including tools and APIs~\citep{mrkl,gorilla,toollm,toolemu}, web, desktop, and workplace environments~\citep{visualwebarena,workarena,weavebench2026}, mobile applications~\citep{appagent,mobileagentv35}, and code or command-line interfaces~\citep{sweagent,terminalbench}. Recent systems combine these modalities by comparing or integrating API and GUI control~\citep{apigui,mcpworld}, augmenting GUI interaction with Python and Bash execution~\citep{coact}, and unifying browser, shell, and extensible tool use~\citep{openclaw}. As these capabilities diversify, multi-agent systems organize specialized agents through role-based interaction and structured workflows~\citep{autogen,agentverse,metagpt,chatdev}; systems such as Magentic-One, UFO, AgentOrchestra, and OWL further separate high-level planning and delegation from specialized execution~\citep{magenticone,ufo,agentorchestra,owl}. Cross-device research extends this organization across distinct device environments: AppCopilot incorporates cross-app and cross-device orchestration on Android devices~\citep{appcopilot}; CRAB supports and evaluates desktop--mobile tasks~\citep{crab}; UFO\textsuperscript{3} coordinates desktops, servers, mobile devices, and edge endpoints through mutable task constellations~\citep{ufo3}; and TopoClaw relies on a primary agent to coordinate planning and execution across devices while accounting for device capabilities~\citep{topoclaw}. OctoNest distinguishes modality feasibility from device feasibility during execution.

\section{Conclusion}

We presented OctoNest, which coordinates device-local modality control and cross-device orchestration to support intra-device and inter-device switching as execution state is revealed through interaction. We introduced CAPEBench to evaluate these capabilities under controlled modality-level, device-level, and mixed perturbations. Experiments show improved execution quality and lower token cost per perfect pass. Our analyses highlight complementary requirements for reliable multi-device execution: device-local control must progressively decompose subtasks and select suitable modalities, while orchestration must maintain and revise the remaining plan using execution feedback. CAPEBench reveals distinct degradation patterns under architectural interventions, exposing limitations obscured by unperturbed evaluation. These findings highlight the importance of coordinating local refinement and global plan revision under partially known execution conditions.




\begingroup
\interlinepenalty=10000
\bibliography{iclr2027_conference}
\endgroup
\bibliographystyle{octonest_references}

\end{document}